\pdfoutput=1

\documentclass[11pt]{article}

\usepackage[preprint]{acl}

\usepackage{times}
\usepackage{latexsym}
\usepackage{graphicx}
\usepackage{tabularx}
\usepackage{booktabs}
\usepackage[linewidth=1pt]{mdframed}
\usepackage{multirow}
\newcolumntype{Y}{>{\centering\arraybackslash}X}
\newcolumntype{C}[1]{>{\centering\arraybackslash}p{#1}}
\newcommand{\tabularxmulticolumncentered}[3] 
    {\multicolumn{#1}
                 {>{\hsize=\dimexpr#1\hsize+\tabcolsep * (2 * (#1 - 1) )+\arrayrulewidth* (#1 - 2)\relax}#2}
                 {#3}}
\usepackage{amsmath}
\usepackage{amsfonts}
\usepackage{relsize}
\usepackage{mathtools}
\DeclareMathOperator*{\argmax}{arg\,max}

\graphicspath{ {./figures/} }

\usepackage[T1]{fontenc}

\usepackage[utf8]{inputenc}

\usepackage{microtype}

\usepackage{inconsolata}

\title{Unraveling and Mitigating Retriever Inconsistencies in Retrieval-Augmented Large Language Models}

\author{Mingda Li$^1$\quad
Xinyu Li$^1$\quad
Yifan Chen$^1$\quad
Wenfeng Xuan$^2$\quad
Weinan Zhang$^1$\thanks{Corresponding author} \\
$^1$\normalsize{Research Center for Social Computing and Information Retrieval}\\[-.05cm]
\normalsize{Harbin Institute of Technology, China}\\[-.05cm]
$^2$\normalsize{XVERSE Technology Inc., China}\\[-.05cm]
{\small\tt\{mdli, xyli, yfchen, wnzhang\}@ir.hit.edu.cn}\\
{\small\tt \{johnxuan\}@xverse.cn}}

\begin{document}
\maketitle
\begin{abstract}
Although Retrieval-Augmented Large Language Models (RALMs) demonstrate their superiority in terms of factuality, they do not consistently outperform the original retrieval-free Language Models (LMs). Our experiments reveal that this example-level performance inconsistency exists not only between retrieval-augmented and retrieval-free LM but also among different retrievers. To understand this phenomenon, we investigate the degeneration behavior of RALMs and theoretically decompose it into four categories. Further analysis based on our decomposition reveals that the innate difference in knowledge sources and the unpredictable degeneration of the reader model contribute most to the inconsistency. Drawing from our analysis, we introduce Ensemble of Retrievers (EoR), a trainable framework that can adaptively retrieve from different knowledge sources and effectively decrease unpredictable reader errors. Our experiments on Open Domain Question Answering show that EoR substantially improves performance over the RALM with a single retriever by considerably reducing inconsistent behaviors.\footnote{Our code and data are available at \url{https://github.com/mingdali6717/Ensemble-of-Retrievers}}

\end{abstract}

\section{Introduction}
Although Large Language Models (LLMs) have shown their superiority in many NLP tasks (\citealp{gpt4, llama2}), they are known to struggle with factual hallucinations (\citealp{mallen2023not, bang2023multitask}) and outdated parametric knowledge (\citealp{dhingra2022time, vu2023freshllms}). Retrieval-Augmented Language Models (RALMs) as an ad-hoc technique have been proven to effectively alleviate these problems (\citealp{ram2023context, vu2023freshllms}). In most RALM systems, a \textbf{retriever} takes the responsibility to retrieve relevant information from some external knowledge sources (e.g., Wikipedia dump \citep{lewis2020retrieval}, search engine \citep{nakano2021webgpt}, parametric knowledge \citep{yu2022generate}) and process them into text chunks to extract the most pertinent content for subsequent generation with a reader model (e.g., filtering, reranking \citep{liu2023webglm}, compression \citep{xu2023recomp}). 

\begin{figure}[t]
    \centering
    \includegraphics[width=\linewidth]{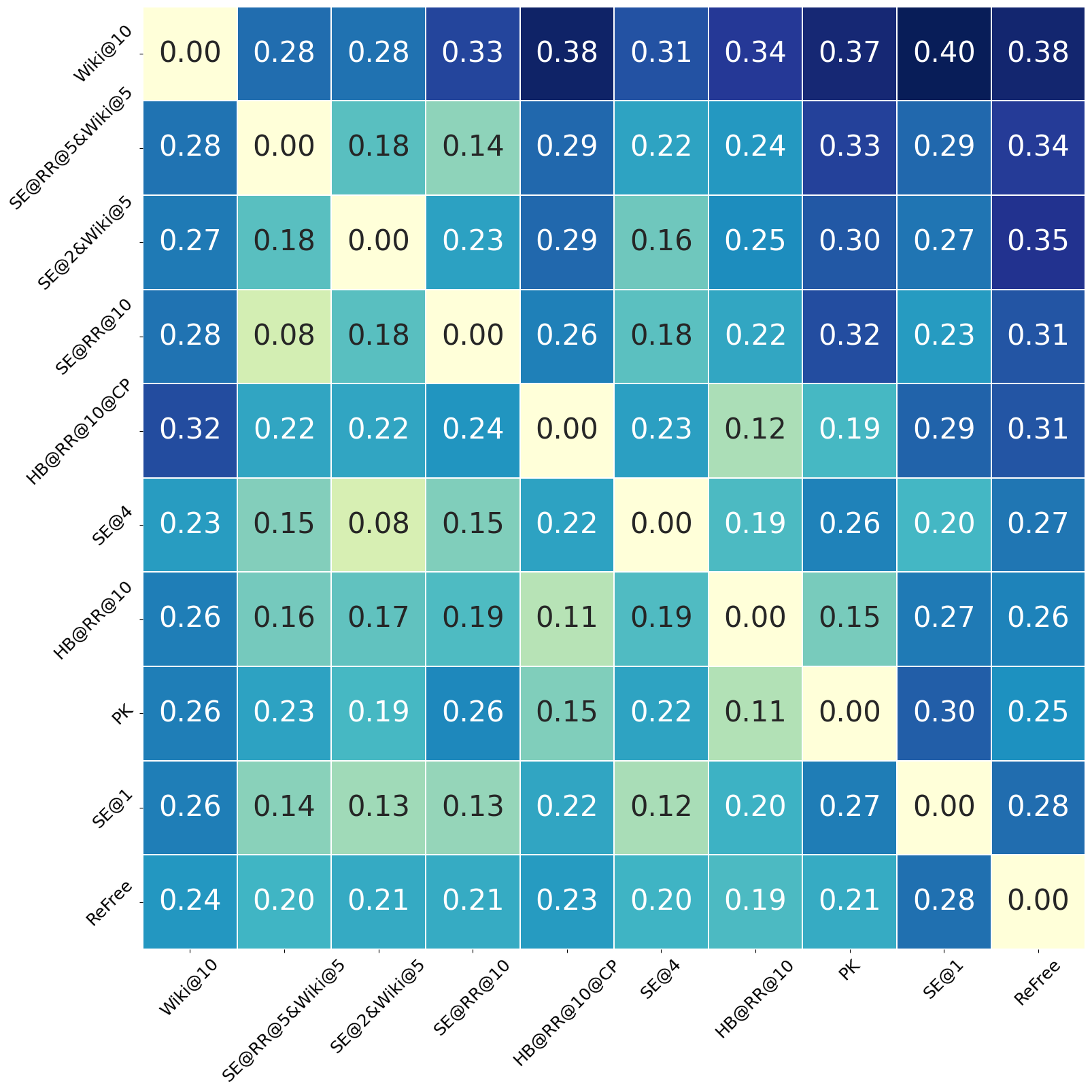}
    \caption{Retriever-to-Retriever Relative Win Ratio heatmap on Natural Questions with ChatGPT as LM. Each cell's number represents the proportion of questions answered incorrectly by the column retriever that was correctly answered by the row retriever. 0 represents all questions correctly answered by the row retriever can be correctly answered by the column retriever, which implies the column retriever consistently outperform the row retriever. See equation \ref{eq1} for formal definition.}
    \label{fig:1}
\end{figure}

Although RALMs show their effectiveness at the corpus level \citep{gao2023survey}, retrieval-augmentation does not consistently promote the original retrieval-free LM but sometimes hurts its performance at the individual example level (\citealp{mallen2023not, yoran2023robust, asai2023selfrag}). In this work, we observe that variability in example-level performance exists not only between retrieval-augmented and retrieval-free LMs, but actually among different retrievers\footnote{We distinguish different retrievers by their knowledge sources and text processing methods. We consider two retrievers equal if and only if their retrieved text chunks are exactly the same for the same query. Retrieval-free could be thought of a singular retriever which reads the query and outputs an empty set.} (e.g. Figure \ref{fig:1}). We call this observed phenomenon as \textbf{retriever inconsistency}. Specifically, we use open-domain question answering (ODQA;\citealp{chen2017odqa}) as our benchmark task and build 15 different retrievers by retrieving from different knowledge sources (search engine, Wikipedia dump and parametric knowledge) and implementing diverse processing methods (truncation, concatenation, reranking and compression). Our experiment reveals that, on average, more than 16$\%$ of questions incorrectly answered by one retriever can be corrected by an alternative retriever, and this result holds for all testing models and datasets. 

To further investigate the reasons behind retriever inconsistency, we theoretically show that RALM's degenerate behaviors on ODQA under regular conditions can be divided into four categories: Retriever Error, Extraction Error, Hallucination Error, and Lucky Guess. We further empirically investigate the example-level error occurrence behaviors of these four types of errors and the results indicate a ubiquitous inconsistent pattern across retrievers for every error type, which collectively contributes to the retriever inconsistency. 

Our further analysis reveals that the innate difference in knowledge sources, such as the absence of post-2018 information in the 2018 Wiki dump or search engine's deficiency in understanding tricky queries, and the inevitable and unpredictable degenerated behavior of the reader model, such as hallucination or the weak robustness to irrelevant context, serve as the main reason of retriever inconsistency.

Inspired by our analysis, we propose Ensemble of Retrievers (EoR), a trainable framework that first samples from RALM with different retrievers and then rejects based on a voting mechanism that measures similarity between answers. Not only can EoR reduce retriever errors by adaptively retrieving from the most appropriate knowledge source, but effectively reduce errors caused by unpredictable degradation of the reader model by comparing answers from different retrievers, based on our observation of inconsistent model error behavior and the intuition that incorrect answers vary while correct answers are always similar. By introducing controlling parameters in our framework, we can easily construct an optimization problem which can be solved by a heuristic search algorithm, and automatically search for the optimal retriever pool used for sampling. Our framework is compatible with any LLM and does not require any training on them. Experiment shows that EoR can effectively improve the performance consistency compared to RALM with a single retriever, thereby improving the corpus performance on ODQA.

\section{Retrievers Are Inconsistent}
To investigate the example-level inconsistent behavior across retrievers and the reasons behind it, we adopt the single-hot short-form ODQA task, which consists of factual questions with short and clear answers, as our benchmark task. The straightforward task format and reliable automatic evaluation metrics \citep{kamalloo2023bem} enable us to quickly and accurately evaluate the correctness of model responses and retrieved documents, facilitating in-depth theoretical and empirical analysis. 
\subsection{Experimental Setup}\label{experiment setup}
\label{sec: experitenal setup}
We adopt the zero-shot in-context RALM \citep{ram2023context} which directly prepends the retrieved documents to the input query based on the prompt template (see Appendix \ref{appendix: prompt templates}). This naive yet efficient framework have been widely used in recent works \citep{gao2023survey}. 
We employ Llama2-chat\textsubscript{7B, 13B} \citep{llama2} and ChatGPT\footnote{\href{https://platform.openai.com/docs/models/gpt-3-5}{gpt-3.5-turbo-instruct}} as our base LM and perform greedy-search on all response generations to reduce the hallucination brought by sampling and guarantee reproducibility.

\textbf{Retrievers}: We characterize retrievers by their individual knowledge sources and the diverse knowledge processing methods. We adopt three different knowledge sources: Search Engine (\textbf{SE}), Wikipedia (\textbf{Wiki}) and model generated parametric knowledge (\textbf{PK}). Specifically, we choose Google as our search engine and directly forward the original query to the Google Search API;\footnote{\url{https://serper.dev/}} We implement DPR \citep{karpukhin2020dpr}, which use English Wikipedia dump from Dec. 20, 2018 as the documents source, to retrieve from Wikipedia; For parametric knowledge, we follow GenRead \citep{yu2022generate} to directly prompt the base LM to generate background documents to answer the query.

As for knowledge processing methods, we adopt four main operations: truncation, concatenation, reranking and compression. Truncation here particularly refer to select top-k text chunks from sorted text list,\footnote{Some knowledge sources embrace innate ranking property such as google and DPR.} denoted by "\textbf{@k}"; Concatenation here specifically refers to concatenate text from different sources, denoted by "\textbf{\&}". Particularly, we use Hybrid (\textbf{HB}) to represent the concatenation of text chunks from all three original knowledge sources; For reranking (denoted by "\textbf{@RR}"), We adopt WebGLM's \citep{liu2023webglm} reranking model, a Contriever \citep{izacard2021contriever} model re-trained with model extracted data. It is reported with better performance than vanilla Contriever; In the case of Compression (denoted by "\textbf{@CP}"), we directly prompt the base LM to summarize the input text, as LLM have demonstrated notable capabilities in extracting information (\citealp{yang2023summarize, liu2023webglm}).

It is intractable to exhaust all retriever combinations, hence we manually design eight typical retrievers (we try to keep their output documents in similar length) and the compressed version of them except for parametric knowledge,\footnote{Model generated documents are generally in short length, hence we do not compress them.} for a total of 15 retrievers. We also regard retrieval-free (denoted by \textbf{ReFree}) as a special singular retriever. Full retrievers list and processing details could be found in Appendix \ref{appendix: retrievers}. We use the combination of operation abbreviations to represent the retriever, the operation priority order follows \textbf{@RR} \textgreater \textbf{@k}\textgreater \textbf{\&} \textgreater \textbf{@CP}. For example, SE@RR@5\&Wiki@5 stands for first reranking the search engine results and then concatenating the top-5 reranked search engine text chunks and the top-5 wiki chunks.

\textbf{Datasets}:\ We experiment on three English ODQA datasets: Natural Questions (NQ; \citealp{kwiatkowski2019nq}), Web Questions (WebQ; \citealp{Berant2013webq}) and TriviaQA \cite{Joshi2017triviaqa}, details refer to Appendix \ref{appendix: datasets}. We evaluate Llama2-chat\textsubscript{7B, 13B} on the full validation split,\footnote{The validation set of WebQ contains only 300 questions, hence we use the train split instead.} whereas for ChatGPT, we randomly sample 500 questions from each split for evaluation because of budget limitation.

\textbf{Evaluation Metrics}: We need two kinds of metrics, one to evaluate the correctness of answers and one to evaluate the example-level inconsistency. Following \citet{kamalloo2023bem}, we adopt BEM score \cite{bulian2022bem}, a semantic similarity metric specifically developed for QA tasks, to evaluate QA accuracy with threshold 0.8.\footnote{We also tried Exact Match, the results are similar.} It is reported to have good correlation with humans and cope with syntactical variation of answers.
\begin{figure}[t]
    \centering
    \includegraphics[width=\linewidth]{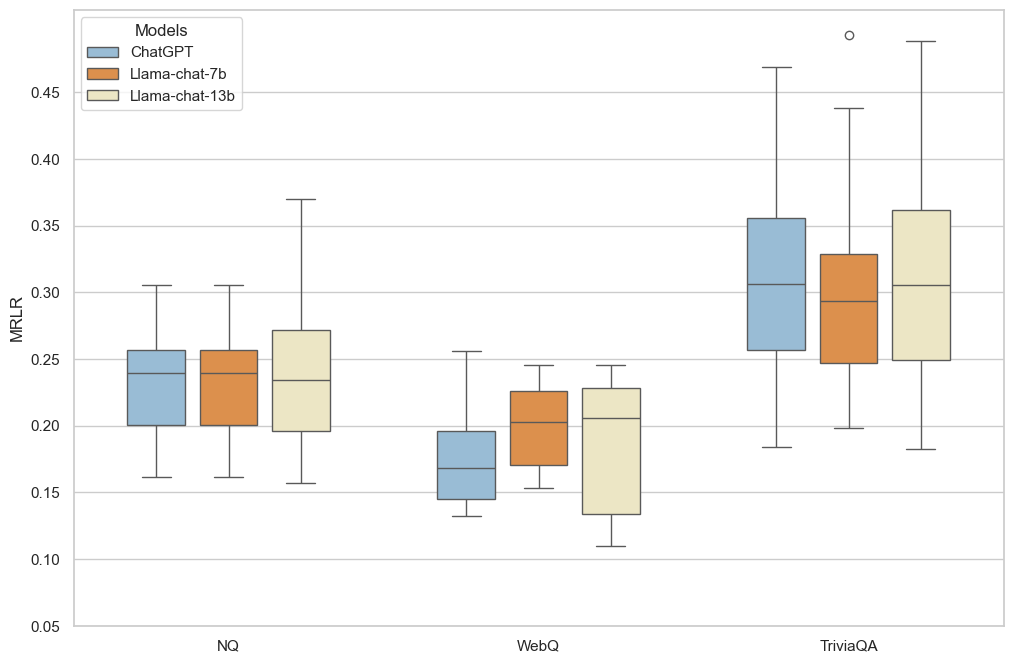}
    \caption{Boxplot displaying the distribution of MRLR of 15 different retrievers across different dataset and models.}
    \label{fig:2}
\end{figure}
\begin{figure*}[t]
    \centering
    \includegraphics[width=\linewidth]{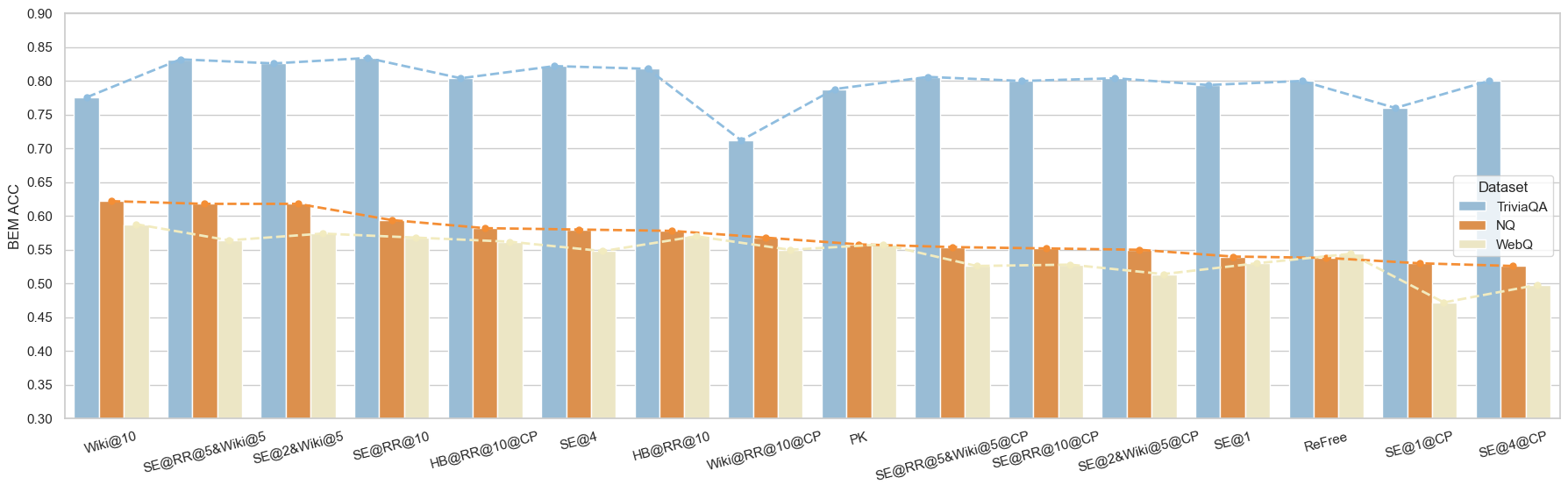}
    \caption{Corpus-level performance of different retrievers evaluated by BEM Accuracy on different datasets with ChatGPT as base LM. The order of retrievers is sorted by performance on NQ.}
    \label{fig:3}
\end{figure*}

As for measuring example-level inconsistency, we propose two naive metrics: Mean Relative Win Ratio (MRWR) and Mean Relative Lose Ratio (MRLR). Assuming we have $M$ different retrievers $\mathcal{R} = \{r_1, r_2, ..., r_M\}$ and a dataset with $N$ samples $\mathcal{D} = \{<q_n, a_n>\}_{n=1}^{N}$. For retriever $r_m$, we can evaluate the correctness of model response for each sample $s_n = <q_n, a_n>$, denoted by $\mathbf{I}^m(n)=1$ if $r_m$ answers correctly on sample $s_n$ otherwise $0$. Then we can calculate the Relative Win Ratio (RWR) of retriever $r_i$ over another retriever $r_j$, which is defined as:
\begin{align}     
    \text{RWR}(i, j)  &= \frac{\sum_{n=1}^N \mathbf{I}^i(n)*(1-\mathbf{I}^j(n))}{\sum_{n=1}^N 1-\mathbf{I}^j(n)} \label{eq1}
\end{align}
Clearly, $\text{RWR}(i,j)$ represents the proportion of questions answered incorrectly by retriever $r_j$ that were correctly answered by retriever $r_i$. The MRWR and MRLR are calculated by respectively averaging RWR across rows and columns:
\begin{align}
    \text{MRWR}(i)  &= \frac{1}{M-1}\sum_{j\neq i} \text{RWR}(i, j)\\  
    \text{MRLR}(i)  &= \frac{1}{M-1}\sum_{j\neq i} \text{RWR}(j, i)
\end{align}
MRLR and MRWR represent the degree of retriever inconsistency. Particularly, MRLR equals zero represents retriever $r_i$ consistently outperforms all other retrievers.

\subsection{Experimental Results}
Figure \ref{fig:1} shows the RWR between different retrievers on NQ with ChatGPT as the base LM. We can observe significant inconsistency  between any two different retrievers. Even for the top-performing retriever, Wiki@10 (62.2$\%$ BEM Acc), 26$\%$ of its failed questions can be correctly answered by the losest-ranked retriever, SE@1 (54.0$\%$ BEM Acc). In Figure \ref{fig:2}, we compare the MRLR of all 15 retrievers across different datasets and models. The result indicates that, on average, more than 16$\%$ of questions incorrectly answered by one retriever can be addressed by an alternative retriever. This phenomenon is prevalent across different base models and datasets, and no evident pattern is observed that larger model can alleviate this phenomenon. The example-level inconsistency also results in the corpus-level performance inconsistency, see Figure \ref{fig:3}, the performance curve of different retrievers on TriviaQA and WebQ do not consistently show a monotonic trend as the sorted NQ curve.

\section{Why Dose Retriever Inconsistency Happen?}
Before delving into the reasons behind retriever inconsistency, we firstly formulate the single-hop short-form ODQA problem and the RALM model. Let $q \in \mathcal{Q}$ denote the factoid question from the ODQA task and $a \in \mathcal{A}$ denote an answer to $q$ (can be correct or wrong). We use $\mathcal{A}_q(a)\subset \mathcal{A}$ to denote the semantic equivalence class of the answer $a$ to the question $q$ and $\mathcal{A}^*_q\subset\mathcal{A}$ for the set of all correct answers to $q$. We assume the existence and uniqueness of $\mathcal{A}^*_q$ for simplicity.\footnote{Existence means $\mathcal{A}^*_q$ is not empty and Uniqueness means that for any $a \in \mathcal{A}^*_q, \mathcal{A}(a) = \mathcal{A}^*_q$. This assumption is reasonable for single-hop short-form ODQA because of its simple task format.}

We consider a vanilla RALM $\mathcal{M}$ consisting of a probabilistic Retriever $\mathcal{R}$ and a probabilistic Reader $\mathcal{G}$. The Retriever $\mathcal{R}$ takes in a query $q$ and return a text string $d \sim \mathcal{R}(q) \in \mathcal{P}(\mathcal{D})$. We call $d$ document and use $\mathcal{D}$ to denote the set of all available documents, $\mathcal{P}(\mathcal{D})$ to denote the set of all probability measures over $\mathcal{D}$. The Reader $\mathcal{G}$ reads the document $d$ and generate a response $y \sim \mathcal{G}(q, d) \in \mathcal{P}(\mathcal{A})$ based on query $q$.\footnote{For deterministic retriever and reader, $\mathcal{R}(q)$ and $\mathcal{G}(q, d)$ collapse to the one-point distribution.} We use random variable $\mathcal{M}(q)$ to represent the answer generated by the RALM and the event, RALM correctly answers the query $q$, can be formally written by $\mathcal{M}(q) \in \mathcal{A}^*_q$. We use $\mathcal{A}_q(a) \in d$ to denote that the document $d$ contains a syntactical variation of the answer $a$ for query q, then we define $\mathcal{D}^*_q = \{d\ | \mathcal{A}_q^* \in d\}$, i.e. the set of documents that contains the correct answer for $q$.

\subsection{Error Decomposition}
\label{sec:decompositon of error}
We now proceed to introduce three key errors contributing to the failures of the RALM.

\textbf{Retriever Error $E_r$}: Given a query $q$ and a retriever $\mathcal{R}$, Retriever Error, denoted by $E_r$, represents the circumstance in which the document returned by Retriever $\mathcal{R}$ does not contain the ground-truth answer for a query $q$, formally defined by:
\[
E_r(q, \mathcal{R}) := \{d \notin \mathcal{D}^*_q\, , \text{given } d \sim \mathcal{R}(q)\}
\]

\textbf{Hallucination Error $E_h$}: Given a query $q$, a document $d$ and a Reader $\mathcal{G}$, Hallucination Error, denoted by $E_h$, stands for the case where the Reader $\mathcal{G}$ generates an answer $y$ that is not present in the document $d$, i.e. 
\[
E_h(q, d, \mathcal{G}) := \{\mathcal{A}(y) \notin d, \text{given } y \sim \mathcal{G}(q, d) \}
\]
which shares a similar definition to the Grounding Error in \citet{DBLP:conf/emnlp/BaekJKPH23}.

\textbf{Extraction Error $E_e$}: Given a query $q$, a document $d$ and a Reader $\mathcal{G}$ , Extraction Error, denoted by $E_e$, stands for the situation where the Reader $\mathcal{G}$ extracts the wrong portion from a correctly retrieved document, formally defined by:
\begin{align}
E_e(q, d, \mathcal{G}) := \{&y \notin \mathcal{A}^*_q \ and\ \mathcal{A}(y) \in d, \notag\\
&\text{given } d \in \mathcal{D}^*_q, y \sim \mathcal{G}(q, d)\}\notag
\end{align}

The probability that these three errors occur for given $q$ and RALM $\mathcal{M}$ can be written by:
\begin{align}
\mathcal{E}_r^{q\text{,} \mathcal{R}} :=& \ \mathbb{P}_{d \sim R(q)}(d \notin \mathcal{D}^*_q) \notag\\
\mathcal{E}_h^{q\text{,} \mathcal{G}}(d) :=&\  \mathbb{P}_{y \sim \mathcal{G}(q, d)}(\mathcal{A}(y) \notin d \, \big|\, d) \notag\\
\mathcal{E}_e^{q\text{,} \mathcal{G}}(d) :=&\  \mathbb{P}_{y \sim \mathcal{G}(q, d)}(y \notin \mathcal{A}^*_q , \mathcal{A}(y) \in d \, \big|\, d\in \mathcal{D}^*_q) \notag
\end{align}

Following above definition, we are able to show that the probability of RALM $\mathcal{M}$ failing on the query $q$, $\mathcal{E}_\mathcal{M}(q) := \mathbb{P}(\mathcal{M}(q)\notin \mathcal{A}^*_q)$, can be decomposed into:\footnote{For deterministic RALM, the decomposition can be written more concisely: $\mathbb{P}(\mathcal{M}(q)\notin \mathcal{A}^*_q)= (1-\mathcal{E}_r)\left(\mathcal{E}_h + \mathcal{E}_e\right) + \mathcal{E}_r\left(1-\mathcal{E}_{luck} \mathcal{E}_{h}\right) $}
\begin{flalign}
    \mathcal{E}_\mathcal{M}(q) &= \mathbb{E}_{d \sim \mathcal{R}(q)}\Big[ \mathbf{I}_{\{d\in \mathcal{D}^*_q\}}\cdot\big(\mathcal{E}^{q\text{,} \mathcal{G}}_h(d)+\mathcal{E}^{q\text{,} \mathcal{G}}_e(d)\big)+ \notag\\
    &\mathbf{I}_{\{d\notin \mathcal{D}^*_q\}} \cdot \big(1-\mathcal{E}^{q\text{,} \mathcal{G}}_{luck}(d) \cdot \mathcal{E}^{q\text{,} \mathcal{G}}_{h}(d)\big)\Big]\label{eq4}
\end{flalign}

where $\mathcal{E}^{q\text{,} \mathcal{G}}_{luck} := \mathbb{P}(y\in \mathcal{A}^*_q \,\big| \, \mathcal{A} (y)\notin d, d\notin \mathcal{D}^*_q)$ represents the probability that $\mathcal{M}$ luckily `hallucinate' the correct answer given an incorrect retrieved document, we call this event \textbf{Lucky Guess} and denote it by $E_{luck}$. Details of the derivation can be found in Appendix \ref{sec:deriviation}.

\subsection{Retriever Inconsistency Stems From Irregular Error Patterns}
\label{sec: error reason discuss}
As shown in equation \ref{eq4}, RALM's failure on the single-hop short-form ODQA problem can be fully described by three errors, Retriever Error $E_r$, Hallucination Error $E_h$ and Extraction Error $E_e$, and a special scenario, Lucky Guess $E_{luck}$. As a result, irregular example-level occurrence of any of these four types of errors\footnote{We still use the term 'error' to represent the Lucky Guess event for simplicity.} will contribute to the inconsistent behavior of the whole RALM. To quantitatively measure the irregular pattern of each error across different retrievers, we follow the similar definition of Relative Win Ratio in section \ref{sec: experitenal setup} to define the RWR for error $E$, $\text{RWR}_{E}$, as: 
\begin{align}     
    \text{RWR}_{E}(i, j)  &= \frac{\sum_{n=1}^N (1-\mathbf{I}_{E}^i(n))*\mathbf{I}_{E}^j(n)}{\sum_{n=1}^N \mathbf{I}_{E}^j(n)} \label{eq:error rwr}
\end{align}
where $\mathbf{I}_{E}^i(n)=1$ if error $E$ occurs for sample $s_n$ and retriever $r_i$, more details refer to Appendix \ref{sec: error analysis}. Therefore, $\text{RWR}_{E}(i, j)$ represents the proportion of Retriever Errors made by retriever $r_j$ that are avoided by $r_i$, and $\text{RWR}_{E}(i, j)=0$ implies that $r_j$ consistently outperform $r_i$.

\begin{figure}[t]
    \centering
    \includegraphics[width=\linewidth]{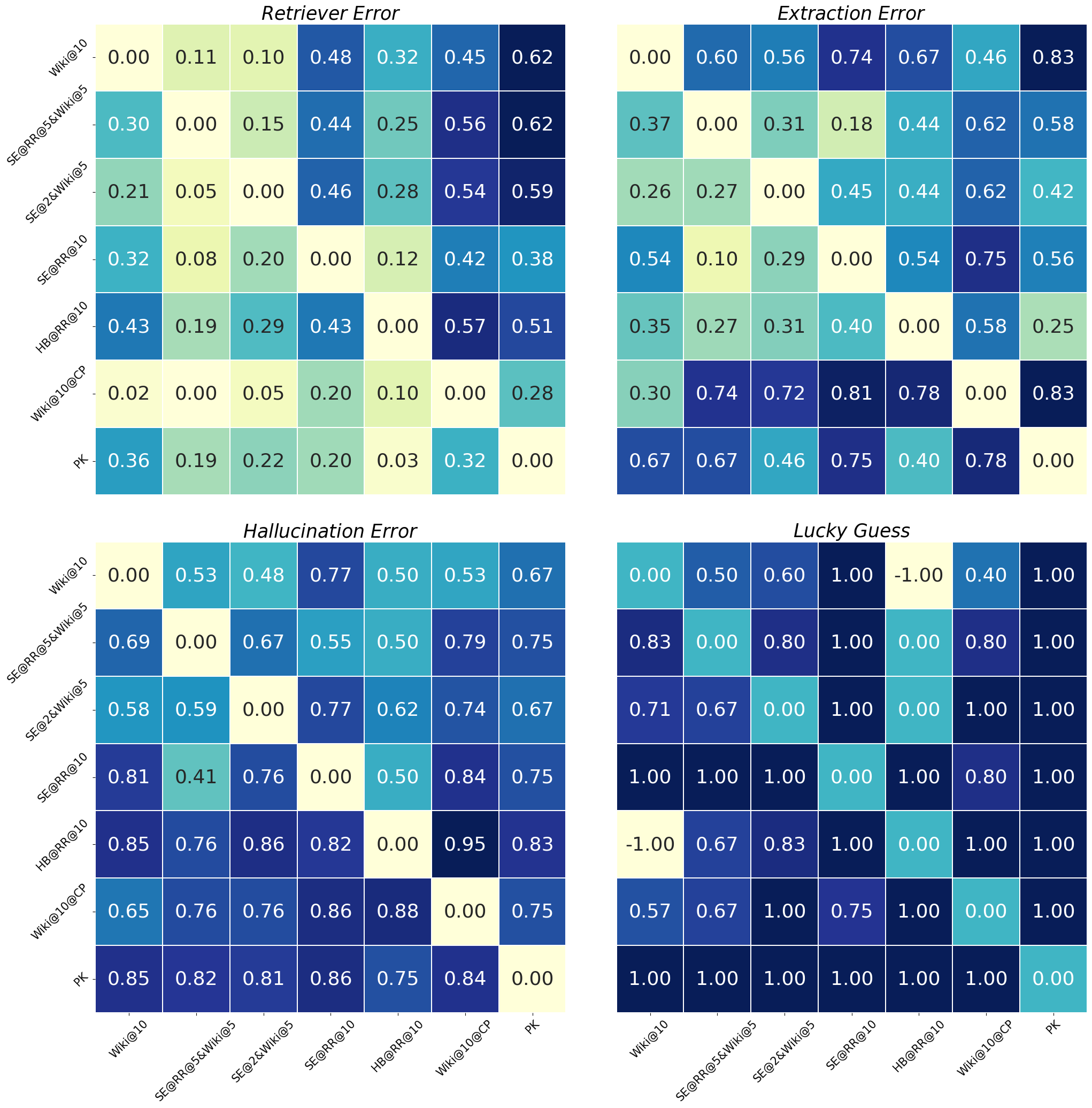}
    \caption{Error Relative Win Ratio between different Retrievers with ChatGPT as base LM, evaluated on NQ validation set. 0 represents the column retriever consistently outperforms the row retriever concerning error occurrence and -1 means at least one of the retrievers is free of this error. We only show part of the result because of space limitation, but the finding is same, more graphs please refer to Appendix \ref{appendix: more figures}.}
    \label{fig:4}
\end{figure}

In Figure \ref{fig:4}, we show the results of different errors, where the retrievers are sorted in descending order by their corpus-level performance (Figure \ref{fig:3}). For Retriever Error, we observe significant bidirectional $\text{RWR}_{E_r}$ between retrievers with different sources (such as 0.62/0.36 for Wiki@10 versus PK) which indicates the innate differences of knowledge sources serve as a main reason for the inconsistency of retrieval errors. As a result, retrievers with hybrid sources (containing concatenation operation "\textbf{\&}") witnessed more consistent behaviors over other retrieves, see the low column $\text{RWR}_{E_r}$ values of them, although still suffer from a small portion of unpredictable errors caused by different processing methods. 

As for Extraction Error, we observe a widespread inconsistency across all retrievers. In particular, even SE@RR@5\&Wiki@5 and SE@2\&wiki@5, which share a large portion of contents and both contain the correct answer,\footnote{This comes from our constructing and Extraction Error estimation methods, see Appendix \ref{sec: error analysis} and \ref{appendix: retrievers}} obtain non-neglectable bidirectional $\text{RWR}_{E_e}$ (0.31/0.27). We believe these ubiquitous inconsistent behaviors stem from Reader $\mathcal{G}$'s weak robustness to long and irrelevant contexts (\citealp{liu2023longcontext, shi2023irrelevantcontext}). A similar phenomenon is observed in Hallucination Error, but with more severe randomness. \citet{kalai2023musthallu} demonstrate that hallucination is inevitable for a statistical reason. 

Therefore, the inconsistent occurrence patterns of three errors collectively contribute to unpredictable RALM's degeneration. Furthermore, Lucky Guess, which compensate the mistakes made by retriever error, also demonstrate an inconsistent behavior and the irregular occurrence will further exacerbating retriever inconsistency.

\section{Ensemble of Retrievers}
\begin{figure}[t]
    \centering
    \includegraphics[width=\linewidth]{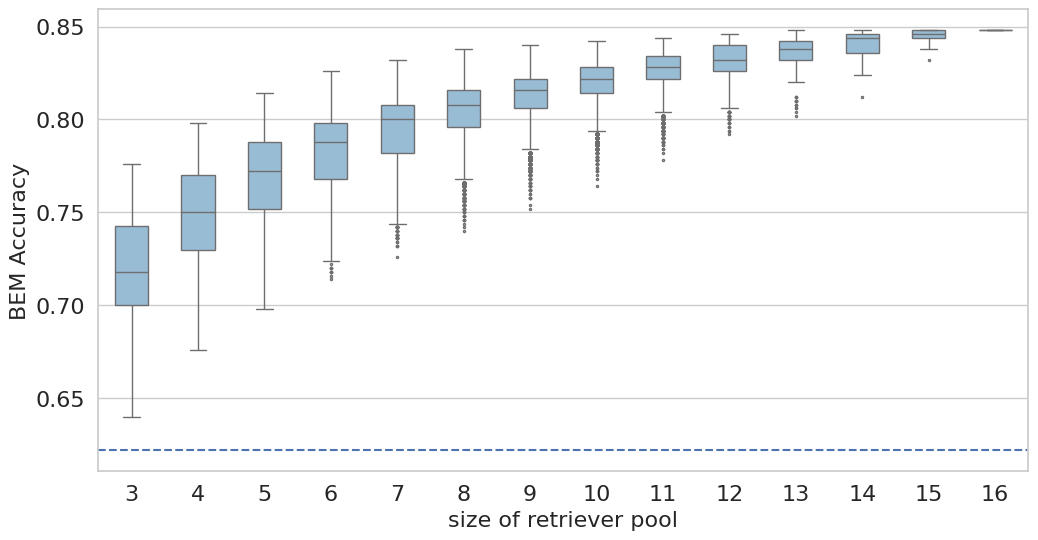}
    \caption{The upper bound of BEM Accuracy by ensembling different retrievers on NQ with ChatGPT as base LM. Each boxplot represents the distribution of the upper bound for different retriever combinations with the same pool size. The dashed line shows the best single retriever performance. }
    \label{fig:5}
\end{figure}

Our analysis provides a strong rationale for adopting an ensemble of retrievers, which can retrieve from different sources, and leverage the irregular error behavior to reduce the degeneration of the reader model. In fact, we can calculate the theoretical upper bound of the ensemble of retrievers by assuming a perfect voting mechanism that will always select the correct answer as long as one retriever outputs the correct one, see figure \ref{fig:5}. It demonstrates an obvious monotonic increasing trend implying the great potential of the ensemble of retrievers. However, the variability in performance within models with the same retriever pool size underscores the importance of understanding both what to include in the ensemble and how to effectively combine them.

\subsection{Our Method}
We propose EoR, a trainable generate-then-rerank framework that can dynamically determine what to retrieve and how to retrieve. Formally, suppose we have $M$ different retrievers $\mathcal{R} = \{r_i\}_{i=1}^{M}$ and a reader model $\mathcal{G}$. EoR accepts an input query $q$ and first generates $M$ responses based on different retrievers, written as $y_m = \mathcal{G}(q, r_m(q)), \ y_m\in \mathcal{A},\  m\in \{1,2,...,M\}$, then the voter module $S_{voter}: \mathcal{A}^M \rightarrow \mathbb{R}^M$ takes in the responses and calculates a score $s_m$ for each response $y_m$, i.e. $\mathcal{S}_{voter}(y_1, y_2,...,y_M)=[s_1, s_2, ..., s_m]$.

The voter module comprises two functions, similarity function $\mathcal{S}_{sim}: \mathcal{A} \times \mathcal{A} \rightarrow \mathbb{R}$ and pooling function $\mathcal{S}_{pool}: \mathbb{R}^{M-1}\rightarrow \mathbb{R}$. The similarity function measures the semantic similarity between two responses. Specifically, we consider a weighted sum of multiple similarity measurement metrics, written as:
\begin{equation}
\mathcal{S}_{sim}(y_m, y_n \big| \omega^s) = \sum_{i}^K \omega^s_i \cdot sim_i(y_m, y_n) \label{eq:6}
\end{equation}
where each $sim(\cdot, \cdot)$ represents a distinct semantic similarity metric, such as EM, BERTScore \citep{zhang2020bertscore}, or the entailment score of a Natural Language Inference (NLI; \citealp{bowman2015nli}) model, and $K$ represents the total number of different metrics. The metric weight $\omega_s$ can be predefined or learned as parameters. As a result, each response $y_m$ corresponds to $M-1$ similarity scores, and the pooling function is responsible for compressing these scores into a single one. Typical pooling functions can be mean, maximum, plurality voting \citep{zhou2012ensemble} or majority voting \citep{wang2023selfconsistency}, detailed formula refer to Appendix \ref{appendix:pooling}. Then the voter score $s_m$ can be formally written as:
\begin{equation}
s_m = \omega_m^r \cdot \mathcal{S}_{pool}(\{\mathcal{S}_{sim}(y_m, y_n\big| \omega^s)\}_{n \neq m}) \label{eq:7}
\end{equation}
where $\omega^r$ is preset or trainable parameters representing the confidence in different retrievers. The final response is selected with the highest score:
\[
\mathcal{M}_{\text{EoR}} (q)= y_{m^*} 
\]
where $m^* = \mathop{\argmax}_m \mathcal{S}_{voter}(\{y_m\}\,\big|\, \omega^s, \omega^r)$
\subsection{Ensemble by Learning}
Like the Stacking methods \citep{wolpert1992stack}, we can use our EoR model to generate data to train the controlling parameters $\omega^s$ and $\omega^r$. Specifically, assuming we have a training dataset $\mathcal{D}_{train} = \{<q_i, a_i>\}_{i=1}^N$. We pass each query $q_i$ through our model $\mathcal{M}_{\text{EoR}}$ and get the corresponding response $y_m^i$ for each retriever and the similarity score $sim_k(y_m^i, y_n^i)$ for each answer pair $y_m^i$, $y_n^i$ under the $k$-th similarity metric. We can further calculate the correctness of each response $y_m^i$ to the ground truth answer $a_i$ with some evaluation metric $g$, written as $g(y^i_m, a_i)$. The evaluation metric $g$ can be EM, BEM, or even human annotation. 

Finding the optimal $\omega^s$ and $\omega^r$ is equivalent to solve the following optimization problem:
\begin{flalign}
&\hspace{2.04em} \mathop{max}\ \ \frac{1}{N}\sum_{i=1}^{N} g(y_{m^*_i}, a_i)\notag\\
&s.t. \ \ \ m^*_i = \mathop{\argmax}_m \,\mathcal{S}_{voter}(\{y_m^i\}\,\big|\, \omega^s, \omega^r)\notag
\end{flalign}
For given $\omega^s$ and $\omega^r$, $\mathcal{S}_{voter}(\{y_m^i\}\,\big|\, \omega^s, \omega^r)$ can be quickly evaluated by equation \ref{eq:6} and \ref{eq:7}. Therefore, we can solve this problem with a heuristic search algorithm, which searches the feasible region by continuously evaluating the objective function. To conduct automatic retriever selection, we can simply replace $\omega^r_m$ with $\omega^r_m\cdot\mathbf{I}_{\omega^r_m > t}$, where $t$ is a hyperparameter. During searching, retrievers with small weights will be directly ignored.

\begin{table*}[t]
\centering
\resizebox{\linewidth}{!}{
\begin{tabularx}{\textwidth}{C{5.5em}  C{4em} Y Y Y Y Y Y Y Y Y}
\toprule
Base &\multirow{2}{*}{$\mathcal{R}$}&\tabularxmulticolumncentered{3}{Y}{NQ}&\tabularxmulticolumncentered{3}{Y}{WebQ} & \tabularxmulticolumncentered{3}{Y}{TriviaQA}\\
Models&&BEM&EM&MRLR&BEM&EM&MRLR&BEM&EM&MRLR\\
\midrule
\multirow{3}{*}{Llama2-chat\textsubscript{7B}}&ReFree&34.35&26.70&23.64&51.82&38.34&24.47&55.57&51.29&48.84\\
&Top $\mathcal{R}$&55.57&46.32&16.33&56.25&43.16&15.66&77.34&72.50&20.11\\
&EoR&\textbf{58.92}&\textbf{50.22}&\textbf{12.36}&\textbf{59.45}&\textbf{46.16}&\textbf{10.66}&\textbf{80.62}&\textbf{75.85}&\textbf{10.62}\\
\hline
\multirow{3}{*}{Llama2-chat\textsubscript{13B}}&ReFree&46.43&35.84&32.84&58.76&44.88&19.44&66.36&60.54&47.92\\
&Top $\mathcal{R}$&62.30&50.94&15.40&\textbf{62.20}&\textbf{49.11}&\textbf{11.48}&82.37&75.99&18.69\\
&EoR&\textbf{64.24}&\textbf{53.07}&\textbf{12.05}&60.63&47.19&12.86&\textbf{83.80}&\textbf{77.77}&\textbf{11.68}\\\hline
\multirow{3}{*}{ChatGPT}&ReFree&52.40&44.60&25.46&59.20&46.00&20.69&81.60&76.8&37.90\\
&Top $\mathcal{R}$&60.20&49.60&16.15&61.00&49.20&18.63&84.40&80.60&23.08\\
&EoR&\textbf{63.00}&\textbf{52.80}&\textbf{12.97}&\textbf{61.60}&\textbf{50.60}&\textbf{13.38}&\textbf{87.40}&\textbf{83.00}&\textbf{12.00}\\
\bottomrule
\end{tabularx}}
\caption{Main results on the test split of NQ, WebQ and TriviaQA. Top $\mathcal{R}$ represents the best-performed single retrieval model on the corresponding test set. \textbf{Bold} number indicates the best performance across retrievers with the same base model and test set.}
\label{table:1}
\end{table*}
\subsection{Experimental Setting}
Same as section \ref{experiment setup}, we use NQ, WebQ and TriviaQA as our experiment datasets and Llama2-chat\textsubscript{7B, 13B} and ChatGPT as our base LM. We search parameters on the validation split (train split for WebQ) and report performance on the test split. For ChatGPT, we randomly sample 500 questions from each split same as Section \ref{experiment setup}. We use BEM accuracy, EM and MRLR as evaluation metrics. 

For EoR, we use the 15 retrievers introduced in section \ref{experiment setup} and ReFree, in a total of 16 retrievers, as our initial retriever pool. We choose EM, BertScore, and NLI\footnote{\url{https://huggingface.co/microsoft/deberta-xlarge-mnli}} as our base similarity metrics in equation \ref{eq:6}, and mean pooling for the pooling function.\footnote{Actually we can search for the optimal pooling methods by transforming to a categorical variable, but our preliminary experiments found that mean pooling is effective enough.} We choose the Nelder-Mead method as the heuristic search method and implement it with SciPy. An upper bound of 0.6 is set for $\omega^s$ and $\omega^r$ to prevent overreliance on a single retriever. We set $\omega^r$ filtering threshold $t$ to 0.1. 

\begin{table}[t]
\centering
\resizebox{\linewidth}{!}{
\begin{tabularx}{\linewidth}{C{9em} Y Y}
\toprule
Retrievers&Train&Test\\
\midrule
Wiki@10&\textbf{62.10}&58.17\\
SE@2\&Wiki@5&61.90&60.97\\
SE@RR@5\&Wiki@5&61.24&\textbf{62.20}\\
EoR&63.02&60.63\\
\bottomrule
\end{tabularx}}
\caption{comparison of Llama2-chat\textsubscript{13B}'s performance with different retrievers on WebQ. \textbf{Bold} number represents the best-performed single retriever on the corresponding split.}
\label{table:2}
\end{table}
\subsection{Results and Analysis}
\textbf{EoR exhibits more consistent behavior than single retriever models}. Table \ref{table:1} presents the results. EoR has a large reduction in MRLR compared to the best-performed single retriever model across all datasets and models except for one exception, which also results in a general corpus-level performance increase. The only exception is Llama2-chat\textsubscript{13B}'s performance on WebQ, which results from the discrepancy in retriever performance between the train and test set, see Table \ref{table:2}. EoR trained on the train set is prone to rely on Wiki@10 which suffers from a significant performance drop in the test set. 

\begin{figure}[t]
    \centering
    \includegraphics[width=\linewidth]{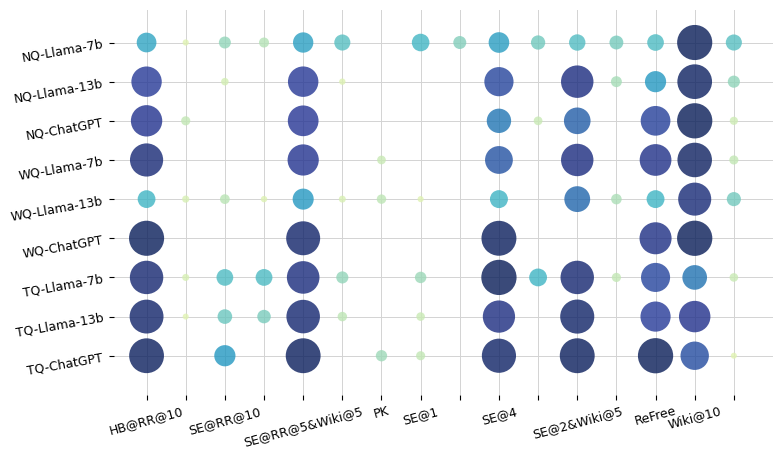}
    \caption{Visualization of Retriever weight $\omega^{r}$ learned with different base LM and datasets. Each row represents the weights from the same EoR model. A larger circle with darker color implies a higher weight on the corresponding retriever.}
    \label{fig:6}
\end{figure}
\textbf{EoR adaptively learns what to retrieve.} Figure \ref{fig:6} visualizes the retriever weights learned by our training methods. Almost every row demonstrates a sparse weight distribution and the remaining retrievers with large weights all performed well on the corresponding training dataset, which implies that EoR can effectively filter out redundant or unreliable retrievers. We can also observe that EoR with Llama-13b trained with WebQ indeed puts most of its weight on Wiki@10, while EoR with Llama-7b and ChatGPT overcome the performance deduction brought by distribution shift by spreading the weight to more retrievers.

\section{Related Work}
\textbf{Degeneration of RALMs}: Many works have empirically demonstrated that retrieval-augmentation sometimes hurts LM's performance (\citealp{ren2023knowledgeboundary, mallen2023not, yoran2023robust}). Some works attribute these failures to incorrect retrieval (\citealp{mallen2023not,chen2022richknowledge}), while some blame the degeneration on the weak robustness of LM on irrelevant or long context (\citealp{ren2023knowledgeboundary,li2023workingmemory,gao2023textcitation}). Particularly, \citet{ren2023knowledgeboundary} found that a large portion of failure cases of ChatGPT come from extracting wrong answers from the supporting documents. \citet{li2023workingmemory} claimed that models are prone to ignore noisy context, although sometimes they can generate correct answer with their parametric knowledge. \citet{liu2023longcontext} showed that LM's performance is vulnerable to the position of relevant information in a long context.

\noindent\textbf{Relieving RALMs's Degeneration}: Some works focus on improving retrieval recall (\citealp{izacard2021contriever,trivedi2023ircot,ma2023queryrewrite}), while some works try to increase RALM's robustness on retrieved context by increasing reader model's ability to leverage context (\citealp{yoran2023robust, liu2023webglm}) or increasing retriever precision (\citealp{xu2023recomp,liu2023webglm}). There are also some works that try to solve this problem from a system perspective, such as using rejection sampling on the reader side to reduce the hallucination (\citealp{menick2022gophercite, asai2023selfrag}), or dynamically deciding whether to retrieve (\citealp{mallen2023not,yoran2023robust,asai2023selfrag, wang2023selfknow,DBLP:journals/corr/abs-2403-14403}). However, all these methods focus on improving the performance of a single retriever-augmented LM, which makes most of them compatible with our EoR framework as long as they can be fitted into a retriever or a reader.
\section{Conclusion}
This work focus on RALM's performance inconsistency across different retrievers. Our results show that retriever inconsistency is ubiquitous across different retrievers and base models. To investigate the reasons behind it, we theoretically decompose RALM's failure into four categories, which serve as a basis for analyzing the degeneration behavior of RALM. Our experiments based on our decomposition reveal that innate differences in knowledge sources and the unpredictable degeneration of the reader model are the main causes for the inconsistency behavior. We further propose Ensemble of Retrievers (EoR), a trainable framework compatible with any LLMs and retrievers. Our experiments demonstrate that EoR can effectively boost RALM's performance by adaptively retrieving from multiple knowledge sources and reducing irregular errors made by the reader model.
\section{Acknowledgements}
We thank the HIT SCIR-DT group members for their
valuable discussions and insightful feedback. This research was supported by the National Key Research and Development Program (No. 2022YFF0902100), Du Xiaoman (Beijing) Science Technology Co., Ltd and Nature Scientific Foundation of Heilongjiang Province (YQ2021F006).
\section{Limitation}
Our theoretical derivation and empirical analysis are all based on the naive singe-hop short-from ODQA setting, whereas multi-hop reasoning and long answer questions are quite common in real-world. We do not choose the task under more complex scenarios because of the absence of reliable automatic evaluation metrics and costly human annotation fees, and we also believe that our conclusion and methods can be generalized to more complex cases. This is because the retriever inconsistency comes from the irregular error occurrence in RALM and intuitively more complex scenario implies weaker robustness. Hence we hypothesize that complex questions will not only exacerbate the errors introduced in section 
 \ref{sec:decompositon of error} but introduce new errors such as reasoning errors, which will result in more severe inconsistent behaviors. But this need further human evaluation to verify. Our EoR framework can also work with more complex questions by using reliable similarity metrics in the voting mechanism and learning by human-annotated data. Example-wise scoring functions (\citealp{asai2023selfrag}) can also be combined with the voter module by some trainable parameters.

Another concern is the computational cost of ensembling multiple retrievers. In fact, EoR can leverage the batch inference of LLMs because of the sharing of the reader model. Hence compared with sampling on the reader side such as self-consistency\citep{wang2023selfconsistency}, the additional computational cost mainly comes from multiple retrievers and the answer similarity calculation with smaller models. If we do not use parametric knowledge from LLMs, most of them are computationally efficient compared to LLMs and can be easily parallelized.

\bibliography{eor}

\appendix

\section{Error Decomposition Derivation}
\label{sec:deriviation}
Given a query $q$ and a RALM model $\mathcal{M}(q) = \mathcal{G}(q, \mathcal{R}(q))$. We consider the probabilistic retriever model $\mathcal{R}(\cdot): \mathcal{Q} \rightarrow \mathcal{P}(\mathcal{D})$, $\mathcal{P}(\mathcal{D})$ denote the set of all probability measures over the set of all documents $\mathcal{D}$, and the probabilistic reader model $\mathcal{G}(\cdot, \cdot): \mathcal{Q} \times \mathcal{D} \rightarrow \mathcal{P}(\mathcal{A})$. $\mathcal{A}$ denote the set of all possible answers. We use $\mathcal{A}^*_q$ to denote the set of all correct answers to $q$ and $\mathcal{A}(a)$ to denote the semantic equivalent class of $a$. We assume the uniqueness of the correct answers in the case of equivalent class, i.e. for any $y_1, y_2 \in \mathcal{A}^*_q$, we have $\mathcal{A}(y_1) = \mathcal{A}(y_2)$, hence $\mathcal{A}^*_q$ is an equivalent class itself.

Now, let's try to decompose the probability that RALM $\mathcal{M}$ answers incorrectly given q.
\begin{flalign}
    &\mathbb{P}(\mathcal{M}(q)\notin \mathcal{A}^*_q)& \notag\\  
    =&\mathbb{E}_{d\sim\mathcal{R}(q)}\Big[\, \mathbb{E}_{y\sim\mathcal{G}(q, d)}\left[ \, \mathbf{I}_{\{y\notin \mathcal{A}^*_q\}}\, \middle| \, d \,\right]\Big]& \notag\\
    =&\mathbb{E}_{d}\Big[\mathbb{E}_{y}\big[\mathbf{I}_{\{y\notin \mathcal{A}^*_q\}} \cdot \mathbf{I}_{\{d\notin \mathcal{D}^*_q\}} + \mathbf{I}_{\{y\notin \mathcal{A}^*_q\}} \cdot \mathbf{I}_{\{d\in \mathcal{D}^*_q\}} \, \big| \, d \,\big] \Big]& \notag\\
    =&\mathbb{E}_{d}\Big[\mathbf{I}_{\{d\in \mathcal{D}^*_q\}}\cdot\mathbb{E}_{y}\big[\mathbf{I}_{\{y\notin \mathcal{A}^*_q\}}  \, \big| \, d\notin \mathcal{D}^*_q\,\big] \Big]+& \notag\\
    &\mathbb{E}_{d}\Big[\mathbf{I}_{\{d\notin \mathcal{D}^*_q\}}\cdot\mathbb{E}_{y}\big[\mathbf{I}_{\{y\notin \mathcal{A}^*_q\}}  \, \big| \, d\in \mathcal{D}^*_q\,\big] \Big]&
\end{flalign}
 $\text{Now, considering}\;\mathbb{E}_{y\sim \mathcal{G}}\left[\mathbf{I}_{\{y\notin \mathcal{A}^*_q\}}  \big|d\notin \mathcal{D}^*_q\right]$:
\begin{flalign}
    &\mathbb{E}_y\big[\, \mathbf{I}_{\{y\notin \mathcal{A}^*_q\}}  \, \big| \, d\notin \mathcal{D}^*_q\,\big]& \notag\\  
    =&\mathbb{P}(y\notin \mathcal{A}^*_q\, \big| \, d\notin \mathcal{D}^*_q)& \notag\\
    =&\mathbb{P}(y\notin \mathcal{A}^*_q\, \big|  \,\mathcal{A} (y)\notin d, d\notin \mathcal{D}^*_q)\cdot\mathbb{P}\left(E_h(q,d,\mathcal{G})\right) +\notag\\
    &\mathbb{P}(y\notin \mathcal{A}^*_q \,\big| \, \mathcal{A} (y)\in d, d\notin \mathcal{D}^*_q)\cdot\left(1-\mathbb{P}\left(E_h\right)\right)\notag\\
    =&\Big[1-\mathbb{P}(y\in \mathcal{A}^*_q \,\big| \, \mathcal{A} (y)\notin d, d\notin \mathcal{D}^*_q)\Big] \cdot \mathcal{E}_{h}(d)+\notag\\
    &1 \cdot \Big(1 - \mathcal{E}_{h}(d)\Big)\notag\\
    =&1-\mathbb{P}(y\in \mathcal{A}^*_q \,\big| \, \mathcal{A} (y)\notin d, d\notin \mathcal{D}^*_q)\cdot\mathcal{E}_{h}(d)\notag\\
    =&1-\mathcal{E}_{luck}(d) \cdot \mathcal{E}_{h}(d)
\end{flalign}
$\mathbb{P}(y\notin \mathcal{A}^*_q \,\big| \, \mathcal{A} (y)\in d, d\notin \mathcal{D}^*_q) = 1$ because if $y \in \mathcal{A}^*_q$, then $\mathcal{A}^*_q \in d$ by $\mathcal{A}(y) \in d$. This is contradicted to the definition of $d\notin \mathcal{D}^*_q$. Actually, if the document $d$ does not contain any semantic variant of the ground-truth answer but contain a semantic variant of $y$, then $y$ can not be a semantic variant of the ground-truth answer.

Similarly,
\begin{flalign}
&\mathbb{E}_{y}\big[\,\mathbf{I}_{\{y\notin \mathcal{A}^*_q\}}  \, \big| \, d\in \mathcal{D}^*_q\,\big]&\notag\\
=&\mathbb{P}(y\notin \mathcal{A}^*_q\, \big|  \,\mathcal{A} (y)\notin d, d\in \mathcal{D}^*_q)\cdot\mathbb{P}\left(E_h(q,d,\mathcal{G})\right) +\notag\\
    &\mathbb{P}(y\notin \mathcal{A}^*_q , \,\mathcal{A} (y)\in d \,\big| \, d\in \mathcal{D}^*_q)\notag\\
=& 1\cdot \mathbb{P}\left(E_h(q,d,\mathcal{G})\right) + \mathbb{P}\left(E_e(q,d,\mathcal{G})\right)\notag\\
=&\mathcal{E}_h(d) + \mathcal{E}_e(d)
\end{flalign}
$\mathbb{P}(y\notin \mathcal{A}^*_q\, \big|  \,\mathcal{A} (y)\notin d, d\in \mathcal{D}^*_q)=1$ because if $y \in \mathcal{A}^*_q$, then $\mathcal{A}(y) = \mathcal{A}^*_q$ by the uniqueness assumption. Then $\mathcal{A}(y) \notin d \Rightarrow \mathcal{A}^*_q \notin d$ which is contradicted to the definition of $d\in \mathcal{D}^*_q$.

Combing equation 4,5 and 6, we get:
\begin{flalign}
    &\mathbb{P}(\mathcal{M}(q)\notin \mathcal{A}^*_q) \notag\\
    =& \mathbb{E}_d\Big[ \mathbf{I}_{\{d\in \mathcal{D}^*_q\}}\cdot\big(\mathcal{E}_h(d) + \mathcal{E}_e(d)\big) \notag\\
    &+ \hspace{0.15cm}\mathbf{I}_{\{d\notin \mathcal{D}^*_q\}} \cdot \big(1-\mathcal{E}_{luck}(d) \cdot \mathcal{E}_{h}(d)\big)\Big]
\end{flalign}

\section{Experiment Setting for Error Analysis}
\label{sec: error analysis}
Assuming we have a dataset with $N$ samples $\mathcal{D} = \{<q_n, a_n>\}_{n=1}^{N}$ and a RALM $\mathcal{M}$ with retriever $\mathcal{R}$. For certain example $<q_n, a_n>$, we can calculate the Answer Correctness Indicator, 
\[
\mathbf{I}^\mathcal{M}_\mathcal{A}(n)=1\ \ if\ \ \mathcal{M}(q_n)=a_n,
\]
the Retriever Error Occurrence Indicator, 
\[
\mathbf{I}^\mathcal{M}_{E_r}(n)=1\ \ if\ \ a_n \notin \mathcal{R}(q_n)
\]
and the Hallucination Error Occurrence Indicator, 
\[
\mathbf{I}^\mathcal{M}_{E_h}(n)=1\ \ if\ \ \mathcal{M}(q_n) \notin \mathcal{R}(q_n)
\]
for each query. We use BEM score with threshold 0.8 to evaluate the answer correctness and Exact Match to evaluate whether a retrieved document contains a certain piece of text, i.e. $a\in d$ if a normalized form of $a$ is matched in $d$. We can then evaluate the occurrence indicator for Extraction Error and Lucky Guess: 
\[
    \mathbf{I}^\mathcal{M}_{E_e}(n) = (1 - \mathbf{I}^\mathcal{M}_{E_r}(n))\cdot(1 - \mathbf{I}^\mathcal{M}_{E_h}(n))\cdot(1-\mathbf{I}^\mathcal{M}_\mathcal{A}(n))
\]
\[
    \mathbf{I}^\mathcal{M}_{E_{luck}}(n) = \mathbf{I}^\mathcal{M}_{E_r}(n)\cdot\mathbf{I}^\mathcal{M}_{E_h}(n)\cdot\mathbf{I}^\mathcal{M}_\mathcal{A}(n)
\]



Because of the training objective of LLM, it tends to generate long answers, which affects the EM accuracy on estimating whether a document contain an answer. Although we have tried several methods to shorten average answer length, such as special instruction, there are still long answers with unimportant content such as "Surely, the answer to the question is". Therefore, before conducting error analysis, we filter out the question with an answer longer than 5 words generated by some retriever and all answers to the same question generated by other retrievers. We denote the filtered dataset as $\mathcal{D}^*$ with size $N^*$.

Following equation  \ref{eq:error rwr}, we calculate the Relative Win Ratio (RWR) for Retriever Error $E_r$ between RALM $\mathcal{M}^i$ and $\mathcal{M}^j$ by:
\begin{align}     
    \text{RWR}_{E_r}(i, j)  &= \frac{\sum_{n=1}^{N^*} (1-\mathbf{I}_{E_r}^{\mathcal{M}_i}(n))*\mathbf{I}_{E_r}^{\mathcal{M}_j}(n)}{\sum_{n=1}^{N^*} \mathbf{I}_{E_r}^{\mathcal{M}_j}(n)} \notag
\end{align}
and the RWR for Hallucination Error $E_h$ by:
\begin{align}     
    \text{RWR}_{E_h}(i, j)  &= \frac{\sum_{n=1}^{N^*} (1-\mathbf{I}_{E_h}^{\mathcal{M}_i}(n))*\mathbf{I}_{E_h}^{\mathcal{M}_j}(n)}{\sum_{n=1}^{N^*} \mathbf{I}_{E_h}^{\mathcal{M}_j}(n)} \notag
\end{align}
As for the Extraction Error $E_e$, notably it is defined under the condition that retriever returns the correct documents. Therefore, when calculating $\text{RWR}_{E_h}(i, j)$, we only consider the circumstance that both retriever in $\mathcal{M}^i$ and $\mathcal{M}^j$ return the correct documents, i.e.
\begin{align}     
    \text{RWR}_{E_e}  &= \frac{\sum_{n=1}^{N^*} (1-\mathbf{I}_{E_e}^{\mathcal{M}_i}(n))*\mathbf{I}_{E_e}^{\mathcal{M}_j}(n)*\mathbf{I}^*_{E_e}(n)}{\sum_{n=1}^{N^*} \mathbf{I}_{E_e}^{\mathcal{M}_j}(n)*\mathbf{I}^*_{E_e}(n)} \notag
\end{align}
where $\mathbf{I}^*_{E_e}(n) = (1-\mathbf{I}_{E_r}^{\mathcal{M}_i}(n))*(1-\mathbf{I}_{E_r}^{\mathcal{M}_j}(n))$.
Similarly,  $\text{RWR}_{E_{luck}}$ is defined by:
\begin{align}     
    \frac{\sum_{n=1}^{N^*} (1-\mathbf{I}_{E_{luck}}^{\mathcal{M}_i}(n))*\mathbf{I}_{E_{luck}}^{\mathcal{M}_j}(n)*\mathbf{I}^*_{E_{luck}}(n)}{\sum_{n=1}^{N^*} \mathbf{I}_{E_{luck}}^{\mathcal{M}_j}(n)*\mathbf{I}^*_{E_{luck}}(n)} \notag
\end{align}
where $\mathbf{I}^*_{E_{luck}}(n) = \mathbf{I}^{\mathcal{M}^i}_{E_r}(n)*\mathbf{I}^{\mathcal{M}^i}_{E_h}(n)*\mathbf{I}^{\mathcal{M}^j}_{E_r}(n)*\mathbf{I}^{\mathcal{M}^j}_{E_h}(n)$.

\section{Implementation Details}
\label{sec: inplementation appendix}
\subsection{Implementation Details of Retrievers}
\label{appendix: retrievers}
Following is the full list of 15 retrievers and corresponding processing methods:\\
\textbf{Wiki@10}: We directly select the top-10 passages returned from DPR (implement with pyserini) and concatenate them. Each passage has an average length of 100 words.\\
\textbf{SE@1}: We fetch and extract contents from the top-1 URL returned by Google API and them truncate it to 1000 words.\\
\textbf{SE@4}: We fetch and extract contents from the top-4 URLs returned by Google API. For the content in each URL, we select 250 words from the beginning including the title. Then concatenate them. \\
\textbf{PK}: We prompt the base LM to generate background documents to answer the query.\\
\textbf{SE@RR@10}: We fetch and extract contents from all URLs returned by Google API. Then we split all contents into chunks with an average of 100 words. We then use the reranking module introduced, a retrained contriever model from WebGLM, to encode the query and each chunk, and select the top-10 chunks with highest cosines similarity.\\
\textbf{SE@2\&Wiki@5}: We concatenate the top-2 chunks from SE@4 and top-5 chunks from Wiki@10.\\
\textbf{SE@RR@5\&Wiki@5}: We concatenate the top-5 chunks from SE@RR@10 and top-5 chunks from Wiki@10.\\
\textbf{HB@RR@10}: Similar to SE@RR@10, we fetch and extract contents from all URLs returned by Google API and then split all contents as long as the document returned by PK into chunks with an average of 100 words. We then put them with the top-20 chunks returned by DPR and use reranking module to rerank all chunks. We select the top-10 as the final result. \\
\textbf{Wiki@10@CP}: We summarize Wiki@10 with our compression model.\\
\textbf{SE@1@CP}: We summarize SE@1 with our compression model.\\
\textbf{SE@4@CP}: We summarize SE@4 with our compression model. \\
\textbf{SE@RR@10@CP}: We summarize SE@RR@10 with our compression model.\\
\textbf{SE@2\&Wiki@5@CP}: We summarize SE@2\&Wiki@5 with our compression model.\\
\textbf{SE@RR@5\&Wiki@5@CP}: We summarize SE@RR@5\&Wiki@5 with our compression model.\\
\textbf{HB@RR@10@CP}: We summarize HB@RR@10 with our compression model. \\

\subsection{Prompt Templates}
\label{appendix: prompt templates}
\textbf{Template for generating Parametric Knoweldge (PK)}:\\
we use \{query\} to represent the placeholder for inserting the corresponding query. This template is following \citep{yu2022generate}.
\begin{mdframed}
Generate a background document to answer the given question.\\
\{query\}. 
\end{mdframed}

\noindent\textbf{Template for Compression (@CP)}:\\
we use \{query\} to represent the placeholder for inserting the corresponding query and \{document\} for the document to be compressed.
\begin{mdframed}
Please truthfully summarize the document below, the summary should contain the most important information relevant to answer the query and be within 200 words:\\
query: \{query\}\\
\\
document: \{document\}\\
\\
summary: 
\end{mdframed}

\noindent\textbf{Template for Retrieval-free Generation}:\\
we use \{query\} to represent the placeholder for inserting the corresponding query. We manually examine many different templates and select the one with highest average validation set performance with our automatic evaluation metrics. \\
Template for ChatGPT:
\begin{mdframed}
Please directly answer the following question within 15 words:\\
\{query\}
\end{mdframed}
Template for Llama2-chat, inspired by \cite{ren2023knowledgeboundary}:
\begin{mdframed}
Please directly answer the following question with one or few words:\\
\{query\}
\end{mdframed}
\hfill\\
\hfill\\
\noindent\textbf{Template for Retrieval-Augmented Generation}:\\
we use \{query\} to represent the placeholder for inserting the corresponding query and \{document\} for the document returned by retriever. \\
Template for ChatGPT:
\begin{mdframed}
Assuming the following paragraphs are true:\\
\\
\{document\}\\
\\
Please directly answer the following question within 15 words:\\
\{query\}
\end{mdframed}
Template for Llama2-chat:
\begin{mdframed}
Assuming the following paragraphs are true:\\
\\
\{document\}\\
\\
Please directly answer the following question with one or few words:\\
\{query\}
\end{mdframed}

\subsection{Datasets}
\label{appendix: datasets}
\textbf{NQ} (\citealp{kwiatkowski2019nq}) consists of questions collected from real Google search queries and the answers are extracted from Wikipedia by humans.\\
\textbf{WebQ} (\citealp{Berant2013webq}) contains questions collected from the Google Suggest API and answers collected by AMT workers based on Freebase.\\
\textbf{TriviaQA} (\citealp{Joshi2017triviaqa}) contains question-answer pairs from several  trivia and quiz-league websites.\\
We use the same dataset splits as GenRead (\citealp{yu2022generate}). They unify the formats of all three datasets and the datasets can be download from this \href{https://drive.google.com/drive/folders/1lFFTklW_0HuR53hLpFdLClgfSAhXn_2f}{URL}. Dataset statistics can be found in Table \ref{table:dataset statistics}.

\subsection{Pooling Functions}
Assuming we have N similarity scores $s_1, s_2, ..., s_N$, then the pooling functions $\mathcal{S}_{pool}: \mathbb{R}^N \rightarrow \mathbb{R}$ are defined as follows:\\
\textbf{Mean Pooling}:
\[
\mathcal{S}_{pool}(s_1, s_2, ..., s_N) = \frac{1}{N}\sum_{i=1}^{N}s_i
\]
\textbf{Max Pooling}:
\label{appendix:pooling}
\[
\mathcal{S}_{pool}(s_1, s_2, ..., s_N) = max\{s_1, s_2, ..., s_N\}
\]
\textbf{Majority Voting}:
\[
\mathcal{S}_{pool}(s_1, s_2, ..., s_N) = \mathlarger{\mathbf{I}}_{\Big\{\sum_i^N \mathbf{I}_{\{s_i > s\}} \,\geq\, \frac{N}{2}\Big\}}
\]
where s is the threshold to identify semantic equivalent answers. The majority voting pooling filters out all answers with the number of semantic equivalent answers less than $\frac{N}{2}$ \\
\textbf{Plurality Voting}:
Assume we have M answers, for the i-th answer, we have calculated M-1 similarity scores $s_{i,1}, s_{i,2}, ..., s_{i,M-1}$. We denote $c^*_i = \sum_j^{M-1} \mathbf{I}_{\{s_{i,j} > s\}}$ which represent the estimated number of semantic equivalent answers given above similarity scores. Then the plurality voting pooling score for the i-th answer is given by:
\[
\mathcal{S}_{pool}(s_{i,1}, s_{i,2}, ..., s_{i,M-1}) = \mathlarger{\mathbf{I}}_{\Big\{c^*_i= max_k\ c^*_k\Big\}}
\]
\begin{table}[t]
\centering
\begin{tabularx}{\linewidth}{Y Y Y Y}
\hline
Datasets&Train&Valid&Test\\\hline
NQ&79,168&8,757&3,610\\
WebQ&3,478&300&2,032\\
TriviaQA&78,785&8,837&11,313\\
\hline
\end{tabularx}
\caption{Dataset statistics}
\label{table:dataset statistics}
\end{table}
\section{Complete Figures}
\label{appendix: more figures}
\begin{figure*}[t]
    \centering
    \includegraphics[width=\linewidth]{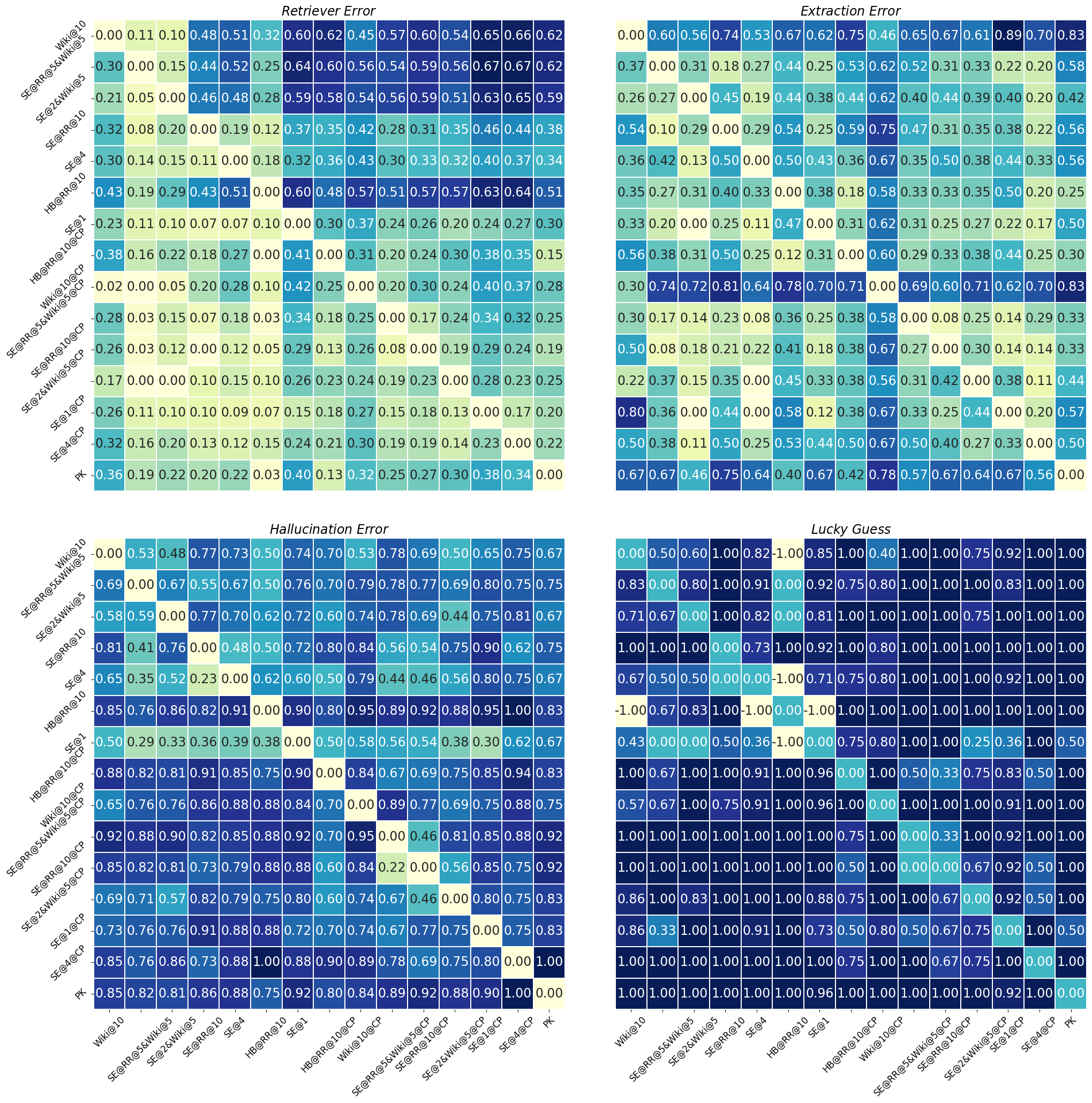}
    \caption{Full Error Relative Win Ratio between different Retrievers with ChatGPT as base LM, evaluated on NQ validation set.}
    \label{fig:nq-full}
\end{figure*}

\begin{figure*}[t]
    \centering
    \includegraphics[width=\linewidth]{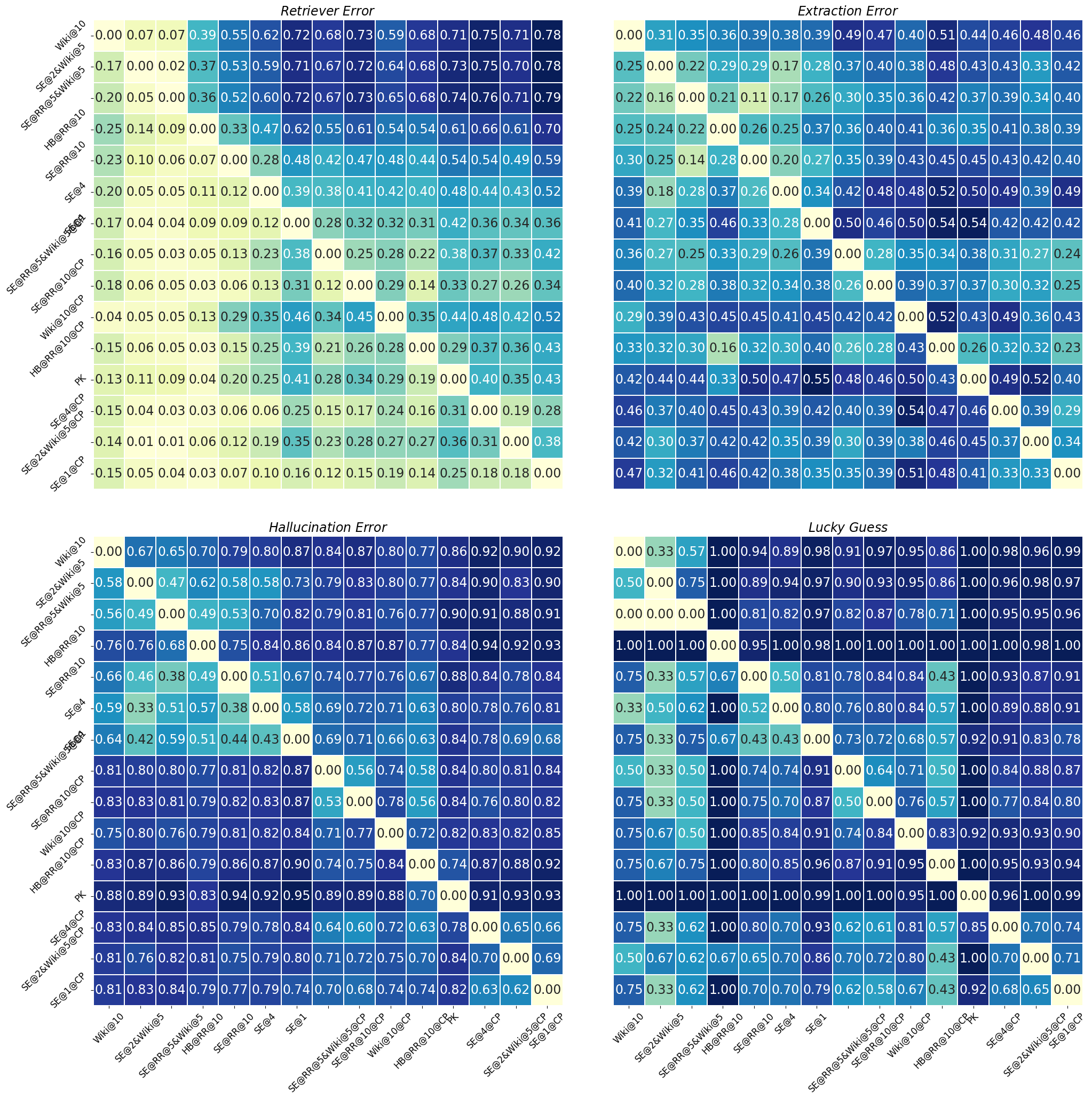}
    \caption{Full Error Relative Win Ratio between different Retrievers with Llama-chat 7b as base LM, evaluated on WebQ train set.}
    \label{fig:wq-full}
\end{figure*}

\begin{figure*}[t]
    \centering
    \includegraphics[width=\linewidth]{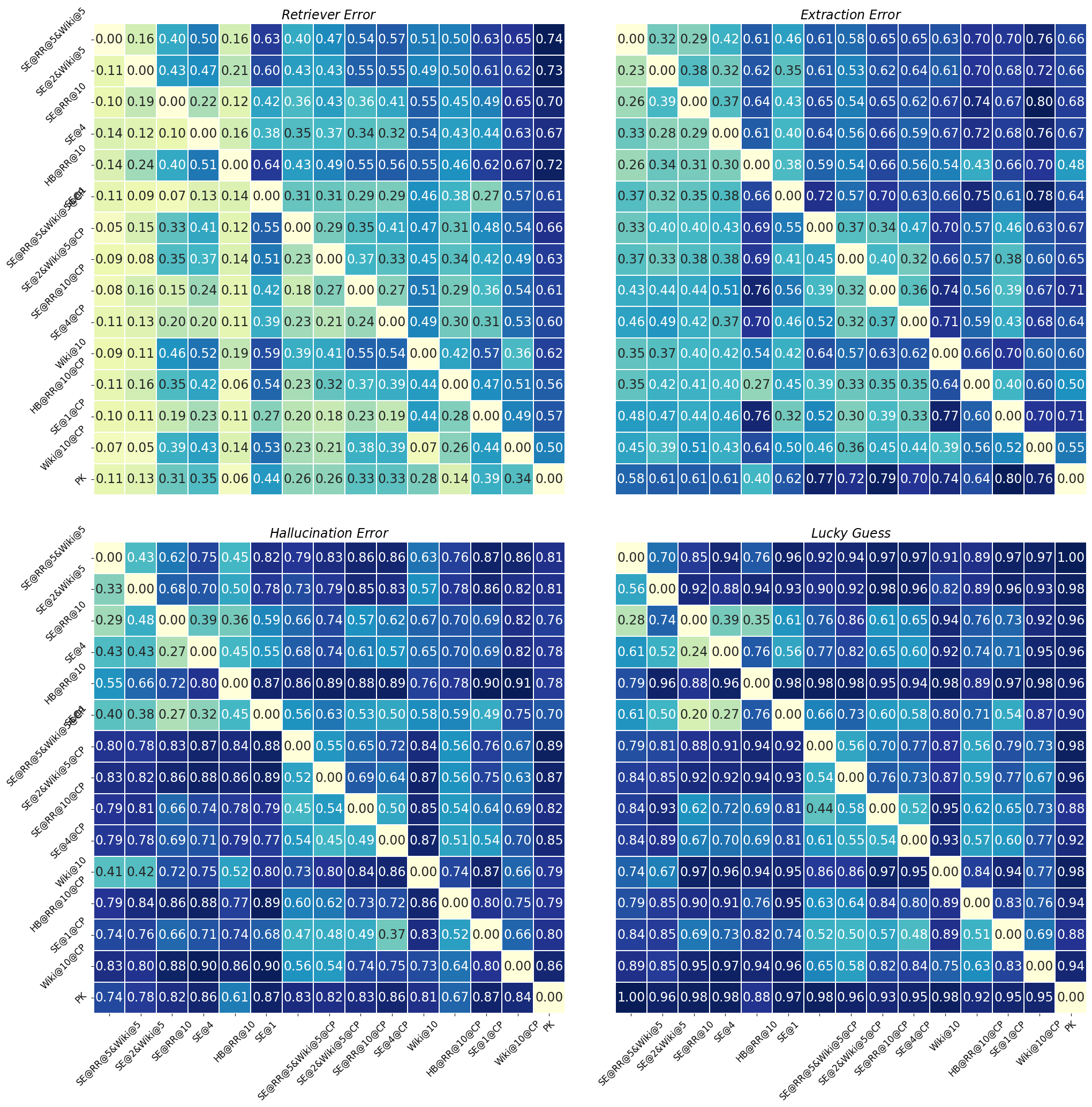}
    \caption{Full Error Relative Win Ratio between different Retrievers with Llama-chat 13b as base LM, evaluated on TriviaQA validation set.}
    \label{fig:tq-full}
\end{figure*}

Figure \ref{fig:nq-full}, \ref{fig:wq-full} and \ref{fig:tq-full} show the Error RWR between all retrievers on different combination of datasets and models. We can see the findings are same as what we discussed in section \ref{sec: error reason discuss}.


\end{document}